\documentclass[letterpaper]{article} 
\usepackage[preprint]{aaai2027}
\usepackage[hyphens]{url}  
\usepackage{graphicx} 
\usepackage{natbib}  
\usepackage{caption} 
\usepackage{booktabs}
\usepackage{amsmath}
\usepackage{tabularx}
\newcommand{\twoline}[2]{\begin{tabular}[t]{@{}r@{}}#1\\#2\end{tabular}}
\newcommand{\twolinel}[2]{\begin{tabular}[b]{@{}r@{}}#1\\#2\end{tabular}}
\newcommand{\twolinec}[2]{\begin{tabular}[b]{@{}c@{}}#1\\#2\end{tabular}}

\title{Pixel Auto-Encoder:\\Reconstructing Language Model Activations Through Pixel Art}
\author{
    Haru Furuichi,
    Phan Xuan Tan
}
\affiliations{
    Shibaura Institute of Technology
}

\begin{document}

\maketitle

\begin{abstract}
Interpreting the high-dimensional internal representations of large language models (LLMs) remains a challenging problem. Recent approaches have explored translating these representations into human-understandable forms, particularly natural language. In this work, we explore the use of images as an alternative target modality for interpreting LLM activations. As a first step toward this goal, we introduce Pixel Auto-Encoder, which converts intermediate LLM activations into 32×32, 32-color pixel art and reconstructs them for subsequent computation within the model. By encouraging the intermediate representations to resemble predefined concept images, we investigate whether LLM activations can be expressed visually while preserving the model's language capabilities. We find that the model can produce target-aligned images for some concepts, and that replacing pixel representations with concept images can influence next-token predictions. However, we also observe a trade-off between visual alignment and language-model performance, as well as sensitivity to pixel-level changes that are difficult to interpret visually. These findings demonstrate the feasibility and limitations of image-based representations of LLM activations.
\end{abstract}

\section{Introduction}

Understanding the internal representations of large language models (LLMs) remains a fundamental challenge in interpretability research. These representations consist of high-dimensional activation vectors that are difficult for humans to interpret directly. Recent approaches have explored translating such representations into human-understandable forms. For example, Natural Language Autoencoders~\cite{NLA} convert LLM activations into natural-language explanations. While natural language provides an intuitive means of describing model representations, images offer an alternative modality for expressing concepts in a form that humans can visually inspect.

The goal of this work is not merely to visualize internal representations, but to reconstruct the original activations from their image-based representations so that the LLM can continue its computation. This introduces two important challenges: whether the intermediate representations can express recognizable images, and whether they can preserve the information required for language generation. A representation optimized for reconstruction may not be visually interpretable, whereas encouraging recognizable images may compromise reconstruction quality.

As a first step toward this goal, we introduce Pixel Auto-Encoder, which converts intermediate LLM activations into 32×32, 32-color pixel art and reconstructs them using an encoder-decoder architecture. The intermediate pixel representations are encouraged to resemble predefined concept images, while the reconstructed activations are passed to subsequent layers of the LLM. We adopt a three-stage training procedure that combines activation reconstruction, image supervision, and downstream adaptation to balance visual representation and language-model performance.

Our experiments show that some intermediate representations resemble their target images, and that replacing pixel representations with concept images can influence subsequent predictions. However, encouraging visual alignment introduces a trade-off with language-model performance, and small pixel-level changes can substantially affect predictions without greatly altering image similarity. These findings demonstrate the potential of image-based representations for interpreting LLM activations while revealing important limitations of the current approach.

\section{Related Work}

Natural Language Autoencoders~\cite{NLA} convert LLM activations into natural-language explanations and reconstruct activations from these explanations. Inspired by this approach, we investigate images as an alternative representation modality.

Our work is also related to Concept Bottleneck Models~\cite{CBM} and Concept Bottleneck Large Language Models~\cite{CBLLM}, which introduce human-interpretable concepts into model computation and enable concept-level interventions. However, concept-based representations may carry information beyond their intended meanings, as discussed by \citeauthor{Mahinpei}~\shortcite{Mahinpei}. We also examine whether the same issue arises in pixel representations supervised by concept images.

\begin{figure*}[t]
\centering
\includegraphics[width=\textwidth]{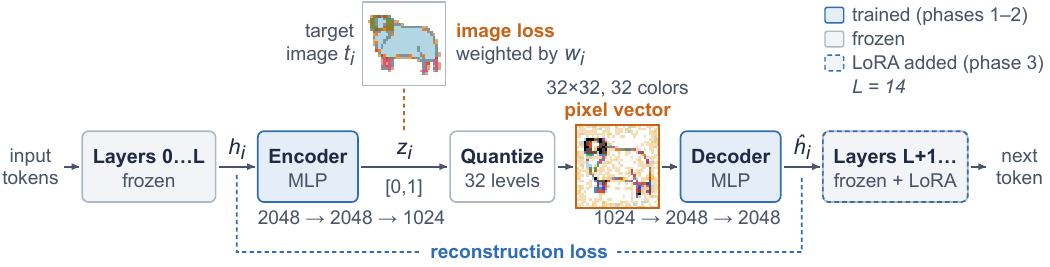}
\caption{Architecture of the Pixel Auto-Encoder. The images are an example at the position of “be” (target image: ram).}\label{fig1}
\end{figure*}

\section{Method}

\subsection{Pixel Auto-Encoder Architecture}

Figure~\ref{fig1} illustrates the Pixel Auto-Encoder, which is inserted between two parts of a pretrained language model.
The Encoder is an MLP that takes as input the residual stream $h_{i}$ of layer $L$ of \mbox{LFM3-1.7B} at position $i$, standardized by its mean and standard deviation, and outputs a 1024-dimensional vector $z_{i}$. Here, $z_{i}$ is kept in the range 0 to 1. The pixel vector is obtained by quantizing each element of this $z_{i}$ into 32 levels. The elements of the pixel vector correspond to the 32×32 pixels, and the 32 levels correspond to a palette ordered by brightness.

However, since the discretized state is not differentiable, gradients are passed through the unquantized $z_{i}$ during training.

The Decoder reads the pixel vector and, like the Encoder, reconstructs the residual stream of layer $L$ with an MLP. Here, the Decoder applies the inverse of the standardization performed in the Encoder and outputs $\hat{h}_{i}$ converted back to the original scale.

\subsection{Concept-to-Image Supervision}

Training requires a target image $t_{i}$ that the pixel vector at each position $i$ should represent, and its weight $w_{i}$. Both are determined using the similarity between $h_{i}$ and a concept vector built for each image.

Concept vectors are built from the residual stream of the same layer as the one used to compute the similarity. Multiple residual streams one token before the token dealing with the target concept are collected, and their average becomes this concept vector. To keep this concept vector from mixing in different usages, we use synthetic text generated by \mbox{LFM3-1.7B} with the word sense specified.

In assigning target images, the similarity between $h_{i}$ at each position $i$ of the training text and all concept vectors is computed. However, so that common components unrelated to the concepts of interest do not affect the similarity, each is whitened beforehand by the mean and covariance of the residual stream. The image of the concept with the highest similarity becomes the target image $t_{i}$ of the pixel vector at position $i$. $t_{i}$ is handled in the same format as the pixel vector.

However, with this method, some concept is forcibly assigned even when a position is only weakly related to every concept. We therefore weighted each position using its similarity to the concept vectors, so that the loss becomes smaller at positions with weak correspondence. The weight $w_{i}$ is determined by the following equation.

\begin{equation}\label{eq-w}
w_i = \operatorname{clip}\!\left(\frac{s_1 - a}{b - a},\, 0,\, 1\right) \times \sigma\!\left(\frac{s_1 - s_2}{\tau}\right)
\end{equation}

\noindent Here, $s_{1}$ and $s_{2}$ are the highest and the second-highest similarity at position $i$, $\sigma$ is the sigmoid function, $a$ and $b$ are the median and the top-1\% value of $s_{1}$ over all positions, and $\tau$ is a constant.

\subsection{Three-Stage Training}

Training is carried out in three phases. In Phase 1, the model is trained with the sole goal of being able to pass through the pixel-vector format without impairing language-model performance. Therefore, the loss here is only the squared error between $h_{i}$ and $\hat{h}_{i}$, each after standardization.

\begin{equation}\label{eq-rec}
\mathcal{L}_{\mathrm{rec}} = \frac{1}{N} \sum_i \frac{\bigl\| \operatorname{norm}(\hat{h}_i) - \operatorname{norm}(h_i) \bigr\|^2}{d}
\end{equation}

\noindent Here, $\mathrm{norm}$ is standardization by mean and standard deviation, $N$ is the number of positions, and $d$ is the dimensionality of the residual stream. Since only the format and the criteria for conversion and reconstruction are learned at this stage, the pixel vector does not become an image that humans can read.

In Phase 2, a loss is added so that the pixel vector can be interpreted as an image. This loss is determined by weighting the error between $z_{i}$ and $t_{i}$ with the weight $w_{i}$ defined in the previous subsection.

\begin{equation}\label{eq-total}
\mathcal{L} = \mathcal{L}_{\mathrm{rec}} + \lambda \cdot \mathcal{L}_{\mathrm{img}}
\end{equation}

\begin{equation}\label{eq-img}
\mathcal{L}_{\mathrm{img}} = \frac{1}{N} \sum_i w_i \cdot \frac{1}{P} \sum_p \bigl( z_{i,p} - c(t_{i,p}) \bigr)^2
\end{equation}

\noindent Here, $P=1{,}024$ is the number of pixels, $z_{i,p}$ is the $p$-th element of $z_{i}$, and $c(t_{i,p})$ is the value $(k+0.5)/32$ corresponding to the color index $k$ of the $p$-th pixel of the target image.

The coefficient $\lambda$ applied to the whole image loss is adjusted automatically using the reconstruction error measured on tuning data not used for training.

Even after the training of Phases 1 and 2, reconstruction error remains and adversely affects language-model performance. Therefore, in Phase 3, the base model up to layer $L$, the Encoder, and the Decoder are frozen, and the layers from layer $L+1$ onward are trained with LoRA so that they can produce outputs close to those of the base model even when they receive $\hat{h}_{i}$. Here, the KL divergence between the next-token probability distributions of the base model and of the model passed through the Pixel Auto-Encoder is used as the loss. This partially restores language-model performance while preserving the quality of the pixel vectors.

\begin{figure}[tb]
\centering
\includegraphics[width=\linewidth]{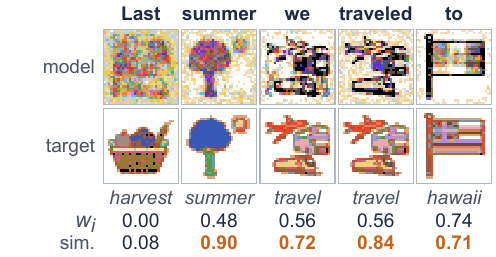}
\caption{Pixel vectors (top) and target images (bottom) at each position of a sentence. sim. is the similarity to the target image.}\label{fig2}
\end{figure}

\section{Experimental Setup}

In our experiments, we used \mbox{LFM3-1.7B}~\cite{LFM3}, a 24-layer language model, and placed the bottleneck at the output of layer 14 ($L=14$). We used Mulberry Symbols~\cite{Mulberry} as the image data, and Wikipedia and LFM3-generated conversations based on Dolly-15k prompts as the training text. Phase 2 was trained for 150,000 steps, and in Phase 3, we applied LoRA to layers 15--23. For evaluation, we used 220 English Wikipedia articles (326,916 tokens) not used for training or tuning. Unless a figure or table states otherwise, the experiments use the model after Phase 3. Full experimental settings are provided in Appendix Table~\ref{tab-setup}.

\section{Experimental Results}

\subsection{Cause of the Decline in Language-Model Performance}\label{sec-cause}

The KL divergence of the model passed through the Pixel Auto-Encoder relative to the base model was 0.392 after Phase 2 and 0.310 after Phase 3. The perplexity after Phase 3 was likewise 24.5, compared with 19.5 for the base model, which shows that it has deteriorated. We conducted an experiment to determine whether this decline in language-model performance is caused by the bottleneck imposed by the architecture or by the image loss. Here, to compare at the point of Phase 2, where the image loss is applied, we continued Phase 1 for 150k steps without adding the image loss (Table~\ref{tab-noimage}).

\begin{table}[tb]
\centering\small\setlength{\tabcolsep}{3pt}
\begin{tabularx}{\linewidth}{@{}>{\raggedright\arraybackslash}X r r r@{}}
\toprule
Model & KL & Perplexity & \twolinel{Recon.}{error} \\
\midrule
Base model & 0 & 19.5 &  \\
No image loss, 150k steps & \twoline{0.118}{(0.106--0.134)} & 20.8 & 0.144 \\
Image loss, 150k steps (ref.) & 0.392 & 26.2 & 0.303 \\
\bottomrule
\end{tabularx}
\caption{Comparison with and without image loss (evaluation data, before Phase 3). Values in parentheses are 95\% bootstrap intervals with articles as the unit.}\label{tab-noimage}
\end{table}

The results show that the degradation caused by passing through the pixel-vector framework is small (KL 0.118), and that most of it is due to the image loss.

\subsection{Similarity between Pixel Vectors and Target Images}\label{sec-sim}

In this subsection, we evaluate how well the pixel vector expresses the assigned target image (Figure~\ref{fig2}). As the metric, we used the cosine similarity between the pixel vector and the target image, $\mathrm{sim}_i = \cos\bigl( c(q_i) - \mu,\, c(t_i) - \mu \bigr)$, where $q_{i}$ is the pixel vector and $\mu$ is the mean of all 1,613 target images. KL is the KL divergence from the base model's next-token distribution to that of the model with the Pixel Auto-Encoder, and the reconstruction error is the value of Eq.~(\ref{eq-rec}).
We also checked whether this similarity changes with the weight of the image loss (Table~\ref{tab-sim}). The overall similarity was 0.224, exceeding the $-0.001$ of the model without the image loss. It was 0.614 at positions whose target-image weight is in the top 10\%, and 0.056 at positions with weight 0.
From this, we found that the model was able to learn according to the target images assigned by the concept vectors and their weights. Hereafter, we use this similarity as the metric for whether the learned image is output at a given position.

\begin{figure}[tb]
\centering
\includegraphics[width=0.97\linewidth]{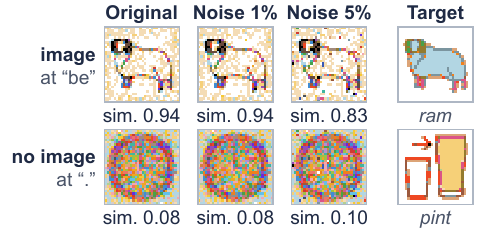}
\caption{Examples with 1\% and 5\% of the pixels changed to random colors. Top: a position where an image appears; bottom: a position where it does not.}\label{fig3}
\end{figure}

\begin{figure*}[t]
\centering
\includegraphics[width=\textwidth]{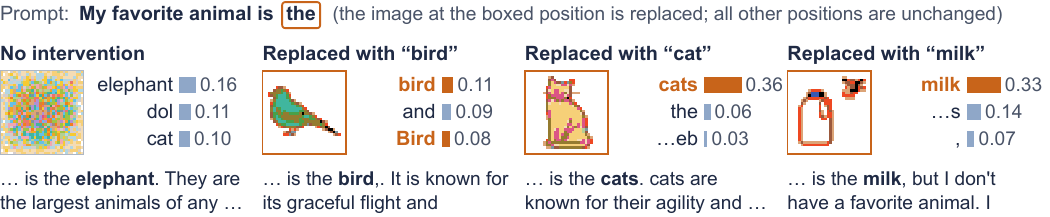}
\caption{Top-3 next tokens and the continuation when the image at the position of “the” is replaced.}\label{fig4}
\end{figure*}

\subsection{Trade-off at Each Position}\label{sec-band}

\begin{table}[tb]
\centering\small\setlength{\tabcolsep}{3pt}
\begin{tabular}{@{}l c c c c c@{}}
\toprule
Weight & \twolinec{Share of}{positions} & \twolinec{Similarity}{to target} & \twolinec{Recon.}{error} & KL & \twolinec{Recon.\ error}{(no image loss)} \\
\midrule
0 & 50\% & 0.056 & 0.214 & 0.242 & 0.132 \\
0--0.04 & 10\% & 0.201 & 0.276 & 0.332 & 0.150 \\
0.04--0.09 & 10\% & 0.283 & 0.309 & 0.376 & 0.154 \\
0.09--0.18 & 10\% & 0.381 & 0.352 & 0.452 & 0.155 \\
0.18--0.34 & 10\% & 0.488 & 0.423 & 0.583 & 0.157 \\
0.34--1 & 10\% & 0.614 & 0.601 & 0.971 & 0.164 \\
Overall & 100\% & 0.224 & 0.303 & 0.392 & 0.144 \\
\bottomrule
\end{tabular}
\caption{Values by target-image weight (before Phase 3).}\label{tab-band}
\end{table}

From Section~\ref{sec-cause}, we know that there is a trade-off between the image loss and KL. We investigated whether this relationship arises not only overall but also at each position. From the similarity between $h_{i}$ at each position of the evaluation data and the concept vectors, we recomputed the weights that would be set during training, and classified the positions where the image loss is strongly applied and those where it is not. For each class, we measured the similarity to the target, the reconstruction error, KL, and so on (Table~\ref{tab-band}).

These results show that, unlike in the model without image loss, whose reconstruction error is almost constant, positions with stronger image supervision tend to exhibit higher reconstruction errors and greater divergence from the base model's predictions. In addition, the noise-like pixel vectors that cannot be recognized as images are thought to be the result of optimization for passing on activations.

\subsection{Noise at Positions Where Images Appear}\label{sec-noise}

Section~\ref{sec-band} shows that positions receiving weak image supervision tend to have low similarity to their target images. We next investigate whether visually aligned pixel representations are also sensitive to small pixel-level changes. We measured the change in KL when applying to the pixel vector operations that cause fine changes small enough not to break the image (Figure~\ref{fig3}). We tested on the positions in the top 10\% of the similarity defined in Section~\ref{sec-sim} (Table~\ref{tab-noise}).

\begin{table}[tb]
\centering\small\setlength{\tabcolsep}{3pt}
\begin{tabularx}{\linewidth}{@{}>{\raggedright\arraybackslash}X r r r@{}}
\toprule
Operation & KL & Perplexity & \twolinel{Similarity}{to target} \\
\midrule
No operation & 0.653 & 41.2 & 0.898 \\
1\% of pixels to random colors & 1.385 & 84.1 & 0.880 \\
5\% & 3.591 & 727 & 0.813 \\
\bottomrule
\end{tabularx}
\caption{Noise applied to the positions in the top 10\% of similarity to the target image.}\label{tab-noise}
\end{table}

These results suggest that downstream predictions depend on pixel-level details that are not fully captured by visual similarity, although the underlying mechanism remains unclear.

\subsection{Intervention by Replacing Images}\label{sec-interv}

Sections~\ref{sec-band} and \ref{sec-noise} suggest that the model also relies on information in the pixel vector that is not captured by its appearance as an image. But does the information of the pixel vector as an image really fail to contribute to language information? To test whether causality can be recognized in the information as an image, we replaced the pixel vector of the LLM partway through its output wholesale with the image of another concept, and examined the change in the next-token output distribution (Figure~\ref{fig4}).

In the example of Figure~\ref{fig4}, the most likely candidate before the replacement was elephant, whereas replacing the image with that of bird or cat made bird or cats the most likely candidate, and, even though the text is about animals, replacing it with milk made milk the most likely candidate.

We conducted 72 trials in total (4 positions × 18 concepts); the word of the replaced concept entered the top 10 in 37 trials and ranked first in 14 (Table~\ref{tab-interv}).

As these numbers show, not every concept behaves as intended, but we found that the image of the pixel vector itself also influences the next-token distribution.

\section{Conclusion}

In this work, we explored the use of images as an alternative modality for interpreting LLM activations and introduced Pixel Auto-Encoder as a first step toward this goal.

Our experiments show that some pixel representations resemble their concept images and that replacing them can influence next-token predictions. However, encouraging visual alignment introduces a trade-off with reconstruction quality and language-model performance. We also observe that small pixel-level changes can substantially affect predictions without greatly changing image similarity, suggesting that visual appearance alone does not fully characterize the information used by the model.

The current approach is further limited by its fixed set of 1,613 concept images. Future work will explore richer visual representations and investigate how faithfully their visual content reflects the information used in subsequent language-model computation.

\bibliography{references}

\appendix
\section*{Appendix}

\begin{table}[!h]
\centering\small
\begin{tabularx}{\linewidth}{@{}>{\raggedright\arraybackslash}X r@{}}
\toprule
 & Similarity \\
\midrule
Overall & 0.224 (0.200--0.252) \\
Positions with target-image weight in the top 10\% & 0.614 \\
Positions with target-image weight 0 (50\% of all) & 0.056 \\
Model without image loss & $-0.001$ \\
\bottomrule
\end{tabularx}
\caption{Similarity between pixel vectors and target images (mean-subtracted cosine). Values in parentheses are 95\% intervals.}\label{tab-sim}
\end{table}

\begin{table}[!h]
\centering\small\setlength{\tabcolsep}{3pt}
\begin{tabularx}{\linewidth}{@{}>{\raggedright\arraybackslash}X c c c@{}}
\toprule
Replaced position & \twolinec{Rank (median)}{before → after} & \twolinec{In}{top 10} & \twolinec{Ranked}{1st} \\
\midrule
She poured herself a cup [of] & 754 → 24 & 8 / 18 & 1 / 18 \\
Last summer we traveled [to] & 1,332 → 8 & 10 / 18 & 3 / 18 \\
My favorite animal is [the] & 823 → 8 & 10 / 18 & 6 / 18 \\
My favorite animal [is] the & 990 → 8 & 9 / 18 & 4 / 18 \\
Total &  & 37 / 72 & 14 / 72 \\
\bottomrule
\end{tabularx}
\caption{Results of replacing the image wholesale with the images of 18 concepts at each of four positions. [ ] marks the replaced position, and the rank is that of the replaced concept's word.}\label{tab-interv}
\end{table}

\begin{table}[!t]
\centering\small
\begin{tabularx}{\linewidth}{@{}>{\raggedright\arraybackslash}p{22mm} >{\raggedright\arraybackslash}X@{}}
\toprule
Base model & \mbox{LFM3-1.7B}~\cite{LFM3} (24 layers, hidden size 2,048) \\
Bottleneck position & Output of layer 14 ($L=14$; layers are counted from 0) \\
Encoder & 2,048 → 2,048 → 1,024 MLP (one hidden layer, GELU) \\
Decoder & 1,024 → 2,048 → 2,048 MLP (one hidden layer, GELU) \\
Trained parameters & About 12.6M in the Encoder and Decoder, about 24.7M in the Phase 3 LoRA \\
Pixel vector & 32×32 pixels, 32 colors (5,120 bits per position) \\
Image data & 1,613 concepts of Mulberry Symbols~\cite{Mulberry} \\
Synthetic text for concept vectors & Generated by \mbox{LFM3-1.7B}. 20 paragraphs per concept (32,260 paragraphs in total). Word senses specified by WordNet~\cite{WordNet} definitions \\
Training text & 1.5M positions of English Wikipedia~\cite{Wikipedia} and 0.5M positions of conversations in which \mbox{LFM3-1.7B} answered the questions of Dolly-15k~\cite{Dolly} \\
Constants of the weight $w$ & $a=0.094$ (median similarity), $b=0.338$ (top 1\%), $\tau=0.01$ \\
Phase 1 & 6,000 steps (8,192 positions per step), AdamW, learning rate 1e-3. English and Japanese text \\
Phase 2 & 150,000 steps (8,192 positions per step), AdamW, learning rate 1e-3 lowered to 30\% with cosine. English only \\
Image-loss coefficient $\lambda$ & Initial value 150, range 75--750, reconstruction-error target 0.33 (measured every 100 steps on 32,768 positions of tuning data) \\
Phase 3 & LoRA (rank 64) on layers 15--23, 8,000 steps, AdamW, learning rate 1e-5. English only \\
Evaluation data & 220 English Wikipedia articles not used for training or tuning (326,916 tokens) \\
Seed & 0 (one) \\
\bottomrule
\end{tabularx}
\caption{Experimental settings.}\label{tab-setup}
\end{table}

\end{document}